\pdfoutput=1

\documentclass[11pt]{article}

\usepackage[final]{acl}

\usepackage{times}
\usepackage{latexsym}
\usepackage{algorithm}
\usepackage{algorithmic}
\usepackage[most]{tcolorbox}
\usepackage{lipsum} 
\usepackage{verbatim} 
\usepackage[T1]{fontenc}

\usepackage[utf8]{inputenc}
\usepackage{listings}
\usepackage{xcolor}

\usepackage{microtype}
\usepackage{booktabs}
\usepackage{xcolor}
\usepackage{amssymb}
\usepackage{amsmath}
\usepackage{inconsolata}

\usepackage{graphicx}
\usepackage{geometry}

\definecolor{darkgreen}{rgb}{0.0, 0.5, 0.0} 

\newcommand{\AgentName}{Table as Thought}

\title{Table as Thought: Exploring Structured Thoughts in LLM Reasoning}

\author{
 \textbf{Zhenjie Sun\textsuperscript{1,2}},
 \textbf{Naihao Deng\textsuperscript{2}},
 \textbf{Haofei Yu\textsuperscript{1}},
 \textbf{Jiaxuan You\textsuperscript{1}}
\\
 \textsuperscript{1}University of Illinois Urbana-Champaign,
 \textsuperscript{2}University of Michigan
\\
}

\begin{document}
\maketitle
\begin{abstract}
Large language models' reasoning abilities benefit from methods that organize their thought processes, such as chain-of-thought prompting, which employs a sequential structure to guide the reasoning process step-by-step.
However, existing approaches focus primarily on organizing the sequence of thoughts, leaving structure in individual thought steps underexplored. 
To address this gap, we propose \AgentName, a framework inspired by cognitive neuroscience theories on human thought.
\AgentName\ organizes reasoning within a tabular schema, where rows represent sequential thought steps and columns capture critical constraints and contextual information to enhance reasoning.
 The reasoning process iteratively populates the table until self-verification ensures completeness and correctness. 
 Our experiments show that \AgentName\ excels in planning tasks and demonstrates a strong potential for enhancing LLM performance in mathematical reasoning compared to unstructured thought baselines. 
This work provides a novel exploration of refining thought representation within LLMs, paving the way for advancements in reasoning and AI cognition.
\end{abstract}

\section{Introduction}
Recent advancements in reasoning have demonstrated that the reasoning capabilities of large language models (LLMs) can be enhanced by introducing structure into the reasoning process \cite{wei2023chainofthoughtpromptingelicitsreasoning, yao2023treethoughtsdeliberateproblem, besta2024got}. For instance, the chain-of-thought approach organizes textual reasoning in a step-by-step manner using a linear chain structure \cite{wei2023chainofthoughtpromptingelicitsreasoning}. Building on this, following works have shown that incorporating more complex organizational structures further improves reasoning performance \cite{besta2024got, yao2023treethoughtsdeliberateproblem}. However, these approaches structure reasoning only at the level of connections between distinct reasoning steps (inter-thought level) and leave the content of individual steps (thought level) unstructured. This raises the critical question: \textit{\textbf{Can LLMs' reasoning abilities be further enhanced by introducing structure  within individual thoughts?}}

To address this question, we draw inspiration from cognitive neuroscience theories of human thought. 
Neuroscientists have found that humans think in a structured way, with the brain's organization facilitating sequential and goal-oriented reasoning. 
\citet{Christoff2000} provided early evidence that the prefrontal cortex supports structured reasoning through a rostrocaudal hierarchy, enabling the processing of increasingly abstract concepts and complex goal-directed behavior. 
Later, \citet{Friston2005}'s predictive coding framework demonstrated how structured cognition emerges from the brain’s ability to build hierarchical models, combining experiences with current input to predict results. 
More recently, Jeff Hawkins \cite{Hawkins2021} argued that humans think in a structured manner, with the neocortex organizing knowledge in certain structures, and thinking arises from neurons activating sequential locations in these frames. 
Building on these insights, we propose investigating whether similarly structured representations can be incorporated into LLMs to enhance their reasoning and planning capabilities.

In this work, we adopt a simple yet effective structural format—a tabular schema—to approximate the structured nature of human thinking processes. 
In our approach, the schema of a table serves as a framework for organizing and navigating knowledge.
Inspired by the sequential processes described in neuroscience—where neurons activate specific patterns step by step \cite{Hawkins2021}—we model these processes as the sequential population of rows in a table, moving across columns according to a predefined schema. 
A single table can encapsulate one or more such structured thought processes, providing a coherent container for organizing and connecting thinking steps and associated information. 
Tables not only represent step-by-step processes for achieving specific goals but also serve as robust frameworks for planning tasks. 
Moreover, utilizing tables as structured representations enables schema design that ensures organization and data integrity, thereby facilitating efficient verification and analysis.

The contributions of our paper are as follows:

\begin{itemize} 
    \item Motivated by insights from cognitive neuroscience regarding the structured nature of human thinking, we propose a novel framework, \AgentName, that injects structure at the thought level. 
    To the best of our knowledge, this is the first exploration and demonstration of integrating structured representations directly into the reasoning process of large language models. 
    \item We demonstrate the advantages of \AgentName\ in tasks requiring planning and mathematical reasoning, highlighting its potential to enhance performance on tasks that demand sequential and goal-oriented thought processes. 
    \item We provide a detailed and comprehensive analysis of \AgentName, offering insights into its functionality and strengths, and comparing the benefits of structured versus unstructured thought representations. 
    We believe these findings can inspire future research into the nature and representation of thought processes in artificial intelligence and computational linguistics. \end{itemize}

\section{Related Work}

\paragraph{Structures in LLM Reasoning} Recent advancements in large language models (LLMs) have increasingly focused on integrating structured processes to enhance reasoning capabilities.
Chain-of-Thought prompting \cite{wei2023chainofthoughtpromptingelicitsreasoning} introduces a step-by-step framework that organizes thoughts in a sequential manner, enabling more coherent reasoning.
Building on this, Tree of Thoughts \cite{yao2023treethoughtsdeliberateproblem} and Graph of Thoughts  \cite{besta2024got} employ hierarchical and networked structures to further enhance problem-solving, leveraging branching and interconnected paths. 
Moreover, self-consistency \cite{wang2023selfconsistencyimproveschainthought} improves reliability by sampling multiple reasoning paths and selecting the most consistent outcome, thereby addressing variability in generated responses.

While these methods excel at organizing reasoning at a macro level—such as through chaining, branching, or aggregating thought paths—they do not address the internal structure of individual thoughts. 
Our work is distinct in that it introduces structure directly at the thought level, refining the granularity of reasoning processes in LLMs. 
By focusing on the internal organization of individual reasoning steps, we provide a novel perspective on enhancing the depth and precision of structured reasoning in LLMs.
\paragraph{Representations of Tables in LLM Inference} Tables have traditionally played a significant role in LLMs for tasks involving the understanding and processing of tabular data, such as knowledge retrieval \cite{cong2024observatorycharacterizingembeddingsrelational}, question answering over structured data \cite{yin-etal-2020-tabert, zhang2024tablellamaopenlargegeneralist}, and tabular reasoning \cite{herzig-etal-2020-tapas, deng-etal-2024-tables}. In these tasks, tables are leveraged only as input for interpretation and manipulation. 

The Chain-of-Table framework \cite{wang2024chainoftableevolvingtablesreasoning} extends the application of tables by employing them as proxies for intermediate thoughts in reasoning tasks involving tabular data. In this framework, LLMs iteratively update a table, forming a dynamic reasoning chain where the table evolves based on intermediate results. While this approach has proven effective on tabular-specific datasets, it remains inherently tied to tasks where tabular data is part of the input or reasoning context.

In contrast, our work redefines the role of tables by utilizing them as a universal framework for structuring and representing the internal thought processes of LLMs in non-table-specific tasks, such as planning and mathematical reasoning. Unlike prior approaches that depend on pre-existing tabular inputs, we employ tables as dynamic containers to organize and manipulate thoughts step by step. This approach enables structured reasoning even in tasks where no tabular data is initially present, bridging the gap between unstructured text-based reasoning and structured problem-solving paradigms. By generalizing the utility of tables beyond table-specific reasoning tasks, our work marks a significant departure from previous methods and demonstrates the versatility of this novel framework.

\begin{figure*}[t] \centering \includegraphics[width=\textwidth]{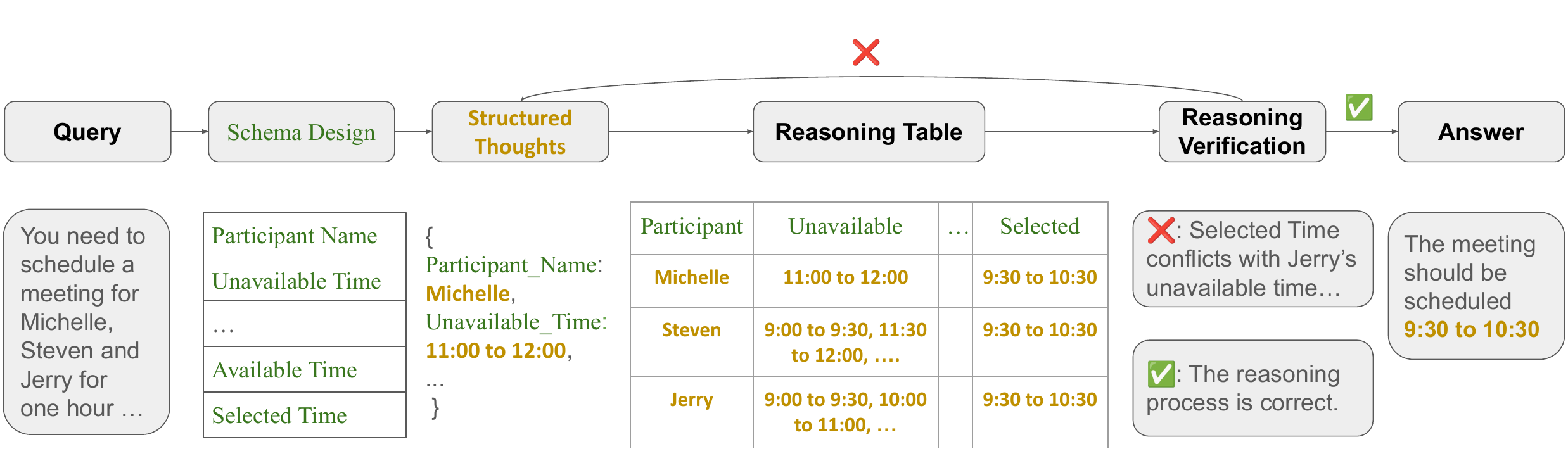} \caption{The Overall Pipeline for Table as Thought Reasoning. The figure illustrates how \AgentName\ structures reasoning by iteratively populating a reasoning table based on the schema, verifying consistency, and updating the table until the final answer is achieved.
} 
\label{fig:table_as_thought} \end{figure*}

\section{\AgentName}

We present the design of the \AgentName~framework, which introduces a novel approach to reasoning in large language models by leveraging tables as structured representations of thoughts.

\paragraph{Table as Thought.} \AgentName\ employs a table as a container to represent one or more structured thoughts. These tables, referred to as \textbf{"reasoning tables"}, encapsulate thoughts and provide a transparent representation of the reasoning process. A reasoning table $T$ is initialized with an original table schema $S$, which is defined by the LLM for a given query $Q$. Structured thoughts $\Theta$ are then generated based on $S$, with each thought corresponding to a row in the reasoning table $T$. The table $T$ is subsequently populated and updated according to these structured thoughts $\Theta$.

The overall reasoning workflow using the reasoning table is illustrated in Figure~\ref{fig:table_as_thought} and formalized in Algorithm~\ref{alg:table_as_thought}.


\begin{algorithm}
\caption{Table as Thought}
\label{alg:table_as_thought}
\begin{algorithmic}[1]
    \REQUIRE Query $Q$
    \ENSURE A table $T$ that satisfies $Q$

    \STATE $S \gets \textsc{DesignSchema}(Q)$ \hfill // Define table schema
    \STATE Initialize an empty table $T$ with schema $S$.
    \WHILE{not \textsc{Sufficient}($T$, $Q$)}
        \STATE $\Theta \gets \textsc{Reflect}(T, Q)$ \hfill // Generate possible updates
        \STATE $T \gets \textsc{UpdateTable}(T, \Theta)$ \hfill // Apply updates if needed
    \ENDWHILE
    \RETURN $T$
\end{algorithmic}
\end{algorithm}

\paragraph{Schema Development Module}
The Schema Development Module dynamically adapts table schemas to accommodate various queries across different reasoning tasks. 
For constraint-planning tasks, where the primary objective is to satisfy constraints, we prompt LLMs to identify the constraints explicitly before designing the schema. 
This ensures that both explicit and implicit constraints are addressed in the reasoning process. 
For mathematical reasoning tasks, the schema is tailored to reflect the logical progression of the reasoning steps, enabling systematic organization of critical information.

The headers in the table schemas are designed to represent essential reasoning steps and key information pertinent to the task. 
These headers act as anchors for organizing and verifying intermediate and final reasoning outputs.

For example, consider the travel planning query:
\begin{quote} \texttt{I plan to travel alone, and my planned budget for the trip is around \$1,100.} \end{quote}
In this case, a key constraint is that the total cost should not exceed \$1,400. 
To address this constraint, the schema must include a header such as \texttt{Cost}, with the type \texttt{Number}, ensuring that the relevant information is captured and evaluated against the budgetary constraint.

For a mathematical reasoning task, such as a question from the GSM8K dataset:
\begin{quote} \texttt{A robe takes 2 bolts of blue fiber and half that much white fiber. How many bolts in total does it take?} \end{quote}
Here, the reasoning process requires consideration of the quantities of blue and white fibers. 
The schema should therefore include keys such as \texttt{Blue Fiber} and \texttt{White Fiber}, ensuring that all relevant elements are systematically tracked and calculated.

\paragraph{Reasoning Verification Module}
The inclusion of this module stems from our findings during experiments that current LLMs sometimes fail to generate the complete reasoning process with structured thoughts to solve a query. 
However, this module is designed not only to verify the completeness of the reasoning process, but also to ensure its correctness.

For constraint reasoning tasks, the module guarantees that all necessary information required to meet the constraints defined in the schema is captured and satisfied. 
Specifically, it verifies whether the constraints identified during the schema development phase are adhered to. 
Constraint checking is generally performed internally by the LLM through reflective reasoning on the generated table, with constraints explicitly listed for verification.  
The structured nature of thoughts in \AgentName\ introduces an additional capability: \textbf{Auto-Check Constraints}, which are constraints set for external verification, performed entirely by the system to ensure that the table adheres to the defined constraints. 
By leveraging the structured representations of \AgentName, Auto-Check Constraints facilitate the systematic validation of intermediate steps and final outputs without relying on the LLM.

For mathematical reasoning tasks, the module evaluates the correctness of the reasoning process by ensuring that the table reflects an accurate and logical reasoning path toward solving the problem. 
This involves checking whether the intermediate and final outputs align with the expected reasoning steps outlined in the schema.

\paragraph{Table Construction Module}
The Table Construction Module iteratively generates structured thoughts and constructs the reasoning table by incorporating the schema and feedback from the reasoning verification module. 
This process involves dynamically adding new thoughts to the table, modifying existing entries, or removing entries that do not align with the schema or query requirements.

The iterative process terminates under one of the following conditions: \begin{enumerate} 
    \item The reasoning table is verified as complete and correct by the reasoning verification module. 
    \item The maximum number of iterations, which is 10 in all our experiments,  is reached. 
\end{enumerate}


\section{Experiments}

\subsection{Tasks and Language Models}
For all tasks, we adopt the original evaluation methods to ensure consistency and comparability.

\paragraph{Constraint Planning Tasks.}
The goal of constraint planning tasks is to generate plans that satisfy both explicit and implicit constraints. We evaluate our approach on two datasets, each presenting different levels of complexity in the expected plans. 
 The TravelPlanner dataset \cite{xie2024travelplannerbenchmarkrealworldplanning} requires LLMs to generate detailed travel plans that adhere to explicit constraints provided in the query, such as budget limitations, as well as implicit constraints derived from common sense. The expected travel plans are highly complex, encompassing multi-day agendas that include transportation, accommodations, and daily attractions. Due to the exceptionally long context required for this task, which results in substantial token costs, we conduct experiments exclusively with GPT-4-o-mini.  
The calendar scheduling task from the NaturalPlan benchmark \cite{zheng2024naturalplanbenchmarkingllms} focuses on generating single-object plans. In this task, LLMs must determine an appropriate meeting time based on explicit constraints, such as the company's working hours and the unavailable time slots of each participant.
\paragraph{Math reasoning tasks.}
We evaluate LLMs using GSM-8K and MATH500 to assess structured mathematical reasoning. GSM-8K \cite{cobbe2021gsm8k} contains 8,000 grade-school-level word problems, testing multi-step reasoning and numerical precision. MATH500 \cite{lightman2023letsverifystepstep} features 500 advanced problems from the MATH dataset \cite{hendrycks2021measuringmathematicalproblemsolving}, covering algebra, calculus, and geometry. It challenges models with tasks requiring symbolic manipulation and deep mathematical understanding. These datasets help evaluate our approach across diverse scenarios, from simple arithmetic to complex problems.

\paragraph{Language Models.}  
The schema design and table construction modules in \AgentName\ require LLMs capable of generating complex, structured outputs that conform to intricate schemas. This capability is natively supported by OpenAI's Structured Outputs Mode, which allows for precise alignment with defined schema requirements. Consequently, our experiments are conducted exclusively on OpenAI's GPT-4-o-mini and GPT-4-o-2024-08-06 models \cite{openai2024gpt4technicalreport}. Expanding the evaluation to include open-source models with similar capabilities remains an area for future work.

\subsection{Text Thought Baselines}  

\paragraph{Direct Prompting.}  
Direct Prompting involves solving queries by directly generating an answer from the input, without prompting for any intermediate reasoning steps. 

\paragraph{CoT Prompting.}  
Chain-of-Thought (CoT) Prompting organizes reasoning as a sequential chain of thoughts, thereby injecting structure into the reasoning process.

\paragraph{Text as Thought.}  This approach differs from \AgentName\ only in its use of unstructured representations for thoughts.
 \textbf{Text as Thought} employs text as the medium for reasoning. This method extends CoT prompting by iteratively updating the reasoning process based on reflection. Each iteration involves generating intermediate reasoning steps, reflecting on their correctness, and refining the reasoning path as needed.
The streamlined process is formalized in Algorithm~\ref{alg:text_as_thought}.

\begin{algorithm}
\caption{Text as Thought}
\label{alg:text_as_thought}
\begin{algorithmic}[1]
    \REQUIRE Query $Q$
    \ENSURE A text $T$ that satisfies $Q$

    \STATE Initialize an empty text $T$.
    \WHILE{not \textsc{Sufficient}($T$, $Q$)}
        \STATE $\Theta \gets \textsc{Reflect}(T, Q)$ \hfill // Generate possible updates
        \STATE $T \gets \textsc{UpdateText}(T, \Theta)$ \hfill // Apply updates if needed
    \ENDWHILE
    \RETURN $T$
\end{algorithmic}
\end{algorithm}

\subsection{Variations of \AgentName}  
To fully explore and understand the boundaries of \AgentName, we introduce two variations to the TravelPlanner task. These variations include \AgentName\ with auto check constraint, which adds complexity to schema design, and \AgentName\ with given schema, which simplifies the task by providing a predefined schema.

\paragraph{\AgentName\ with Auto-Check Constraint.}  
This variation builds on the vanilla \AgentName\ by requiring the LLM to add additional constraints during schema design to ensure data integrity and reflect the constraints present in the query. For instance, if a TravelPlanner query includes budget constraints, the LLM is expected to design a schema with headers like \texttt{Cost} and enforce a rule ensuring that the sum of the column does not exceed the specified budget. By introducing this variation, we aim to explore the boundaries of LLMs in designing complex reasoning structures and handling intricate schema requirements.  

\paragraph{\AgentName\ with Given Schema.}  
In this variation, the LLM is provided with a predefined schema, as shown in Table~\ref{tab:given_vs_LLM_schema}, rather than designing the schema independently. The given schema is derived from the evaluation pipeline of the TravelPlanner task \cite{xie2024travelplannerbenchmarkrealworldplanning}, where answers are processed into tables following this schema before evaluation. This variation serves as a comparative baseline to assess the effectiveness and adaptability of schemas designed by LLMs compared to fixed, predefined schemas.

\begin{table*}[h]
\centering
\resizebox{\textwidth}{!}{
\begin{tabular}{lccc|ccc}
\toprule
\textbf{Metric} & \textbf{Direct} & \textbf{CoT} & \textbf{Text as Thought} &  \multicolumn{3}{c}{\textbf{\AgentName}} \\
& & & & \textbf{Vanilla} & \textbf{w/ Auto-Check  constraint} & \textbf{w/ Given Schema} \\
\midrule
Delivery Rate ($\%$) & 100.0 & 100.0 & 100.0 & 100.0 & 99.4 & \textbf{100.0} \\
Commonsense Constraint Micro Pass Rate ($\%$) & 68.3 & 69.0 & 68.3 & 64.4 & 63.8 & \textbf{70.1} \\
Commonsense Constraint Macro Pass Rate ($\%$) & 2.22 & 2.22 & 0.556 & 0.0 & 0.0 & \textbf{3.33} \\
Hard Constraint Micro Pass Rate ($\%$) & \textbf{7.62} & 6.19 & 3.81 & 3.33 & 1.90 & 5.95 \\
Hard Constraint Macro Pass Rate ($\%$) & 4.44 & 4.44 & 2.78 & 1.67 & 0.556 & \textbf{5.00} \\
Final Pass Rate ($\%$) & 0.556 & 0.556 & 0.0 & 0.0 & 0.0 & \textbf{1.11} \\
\bottomrule
\end{tabular}
}
\caption{Evaluation results for different models and prompt methods on TraverPlanner Tasks on GPT4o-mini}
\label{tab:travel_planner_transposed_split}
\end{table*}

\begin{table}[th!]
    \centering
    \resizebox{0.5\textwidth}{!}{
        \begin{tabular}{lcccc}
            \toprule
            & Direct & CoT & Text as Thought & \AgentName \\
            \midrule
            GPT-4o & 64.0 & 64.5 & 69.4 & \textbf{74.8} \\
            GPT-4o-mini & 36.2 & 38.4 & 38.4 & \textbf{42.3} \\
            \bottomrule
        \end{tabular}
    }
    \caption{Performance of GPT-4o and GPT-4o-mini models under different prompting methods for calendar scheduling.}
    \label{tab:calendar_scheduling_results}
\end{table}
\begin{table}[h!]
    \centering
    \resizebox{0.5\textwidth}{!}{
        \begin{tabular}{lcccc}
            \toprule
            & Direct & CoT & Text as Thought & \AgentName \\
            \midrule
            \multicolumn{5}{c}{\textbf{MATH500}} \\
            GPT-4o & 75.0 & 72.2 & 72.6 & 64.2 \\
            GPT-4o-mini & 65.4 & 65.2 & 63.4 & 47.8 \\
            \midrule
            \multicolumn{5}{c}{\textbf{GSM8K}} \\
            GPT-4o & 95.4 & 95.9 & 95.7 & 94.1 \\
            GPT-4o-mini & 93.9 & 93.6 & 92.9 & 92.4 \\
            \bottomrule
        \end{tabular}
    }
    \caption{Performance of GPT-4o and GPT-4o-mini models under different prompting methods for MATH500 and GSM8K.}
    \label{tab:combined_results}
\end{table}

\begin{table}[h!]
    \centering
    \resizebox{0.5\textwidth}{!}{
        \begin{tabular}{lccc}
            \toprule
            & Direct & CoT & Text as Thought  \\
            \midrule
            \multicolumn{4}{c}{\textbf{MATH500}} \\
            GPT-4o & 4.4/25.0 & 5.4/27.8 & 4.4/27.4 \\
            GPT-4o-mini & 2.0/36.6 & 2.4/34.6 & 2.8/34.8 \\
            \midrule
            \multicolumn{4}{c}{\textbf{GSM8K}} \\
            GPT-4o & 1.59/4.62 & 1.29/4.09 & 1.60/4.33  \\
            GPT-4o-mini & 1.59/6.14 & 2.12/6.37 & 2.50/7.13  \\
            \bottomrule
        \end{tabular}
}
    \caption{The Percentage of Questions that Table as Thought successfully work out while other prompting methods failed vs failed rate of other prompting methods  }
    \label{tab:table_sucess_text_failed_math}
\end{table}

\section{Results}

\subsection{Calendar Scheduling Task}  
\AgentName\ achieves the highest performance among all prompting methods on the Calendar Scheduling Task, as shown in Table~\ref{tab:calendar_scheduling_results}. On GPT-4o, \AgentName\ improves performance by 10.8\% over Direct Prompting and achieves a 5.4\% improvement compared to the Text as Thought baseline. This highlights the advantage of using tables as structured representations for planning over unstructured text-based representations. A similar trend is observed with GPT-4o-mini, where \AgentName\ outperforms other methods, suggesting the robustness of table-based reasoning for simpler constraint reasoning tasks like calendar scheduling.  

For GPT-4o, the improvement from Direct Prompting to CoT Prompting is minimal (0.5\%), indicating that chain-like reasoning structures may already be embedded in the model's reasoning process. However, incorporating self-verification through Text as Thought yields a 4.9\% improvement. Importantly, transitioning from unstructured thoughts to structured tables results in a more substantial performance boost (5.4\%), underscoring the benefits of structured representations in reasoning tasks.

For GPT-4o-mini, a less advanced model, CoT Prompting achieves a moderate 2.2\% improvement over Direct Prompting, but Text as Thought fails to provide any additional gains. This suggests that GPT-4o-mini lacks both the natural incorporation of chain-like structures in its reasoning and the self-verification capability to improve performance on text-based reasoning tasks. In contrast, \AgentName\ demonstrates a significant 4.4\% improvement over CoT Prompting, reinforcing the effectiveness of introducing structure at the thought level over chain-like structures at the reasoning level for less advanced models.

\subsection{TravelPlanner Task}
Table~\ref{tab:travel_planner_transposed_split} shows that \AgentName\ with a given schema achieves the best performance across most metrics in the TravelPlanner task, underscoring the potential of structured thoughts. The significant improvement from vanilla \AgentName\ to \AgentName\ with a given schema highlights that current LLMs struggle to design effective table schemas for achieving complex objectives. This limitation will be analyzed in more detail in the next section.

The results reveal an important trend: on a challenging task like TravelPlanner, which demands reasoning toward a complex objective, introducing increasingly sophisticated structures into the reasoning process can lead to performance degradation. Specifically, methods that incorporate additional complexity—such as chain-of-thought (CoT) prompting, self-reflection in Text as Thought, and rule-constrained structured thoughts in \AgentName\ with Auto-Check constraint—tend to perform worse compared to simpler approaches. The exception is \AgentName\ with a given schema, which avoids this degradation by relieving the LLM of the need to design its own schema, allowing it to focus solely on reasoning within a predefined structure.

\subsection{Math Reasoning Tasks}

Table~\ref{tab:combined_results} highlights a general trend in the MATH500 and GSM8K tasks: introducing additional complexity into the reasoning process often leads to a performance drop, particularly for less capable models like GPT-4o-mini. For instance, on MATH500, the performance of both GPT-4o and GPT-4o-mini decreases as the reasoning structures become more sophisticated, from Direct Prompting to Text as Thought to \AgentName. This effect is especially pronounced for GPT-4o-mini, where the performance of \AgentName\ falls to 47.8\%, compared to 65.4\% with Direct Prompting. A similar trend is observed on GSM8K, where the addition of more structured reasoning methods results in marginal performance degradation. These results suggest that LLMs may already be overfitted to math reasoning tasks, as noted in recent studies \cite{mirzadeh2024gsmsymbolicunderstandinglimitationsmathematical, zhang2024carefulexaminationlargelanguage}.

Despite this general trend, \AgentName\ demonstrates its potential to improve performance by successfully solving questions that text-thought-based methods fail to address, particularly with more capable models like GPT-4o. Table~\ref{tab:table_sucess_text_failed_math} provides a detailed breakdown of the percentage of questions that \AgentName\ solves, which were missed by other methods. On MATH500, \AgentName\ resolves approximately 20\% of such questions, while on GSM8K, this figure exceeds 30\%. These findings underscore the utility of structured reasoning in identifying alternative pathways to solutions that text-based reasoning methods may overlook.

\section{Analysis}
\subsection{Effect of Schema Design on Reasoning Structures}
Schema design plays a pivotal role in structuring the reasoning paths of Calendar Scheduling tasks. Different schemas determine the granularity of the reasoning process, which in turn affects model performance. 

Table \ref{tab:multi_one_schema} shows that in the \textbf{one-row schema}, the reasoning process is concise: the LLM identifies all available time slots for participants in a single step and selects a suitable meeting time. This schema produces a single-row table, encapsulating the reasoning process in a compact form. In contrast, the \textbf{multi-row schema} divides the process into finer-grained steps. The LLM first extracts unavailable and preferred time slots for each participant. It then computes available time slots before aggregating this information to finalize the meeting time. This approach results in a table with multiple rows, each representing an intermediate reasoning step, and provides a more detailed reasoning path.

Table~\ref{tab:multi_vs_one} shows that schema complexity impacts performance differently for advanced and less capable models. For GPT-4o, the multi-row schema outperforms the one-row schema, achieving 80.28\% accuracy compared to 72.93\%. This suggests that the finer-grained reasoning path introduced by the multi-row schema aligns well with GPT-4o’s stronger table reasoning capabilities. By explicitly structuring intermediate steps, the multi-row schema allows GPT-4o to better manage constraints and ensure reasoning correctness. On the contrary, GPT-4o-mini performs better with the simpler one-row schema (45.05\% vs. 43.46\% for the multi-row schema). This indicates that the increased complexity of the multi-row schema exceeds the model’s table reasoning and verification abilities, leading to performance degradation. 
\begin{table}[h]
    \centering
    \resizebox{0.5\textwidth}{!}{

    \begin{tabular}{lc}
        \toprule
        & \textbf{Schema Example} \\
        \midrule
        One Row  & Time Slot, Jesse Availability \\
        & Kathryn Availability, Megan Availability \\
        & All Participants Available, Earliest Availability \\
        \midrule
        Multi Row  & Participant Name, Availability Start Time \\
        & Availability End Time, Meeting Duration \\
        & Work Hours Constraint, Schedule Constraint \\
        & Preference Constraint, Proposed Meeting Time \\
        \bottomrule
    \end{tabular}
    }
    \caption{Schema examples for Multi Row Thought and One Row Thought}
    \label{tab:multi_one_schema}
\end{table}

\begin{table}[h]
    \centering
    \begin{tabular}{lcc}
        \toprule
        & \textbf{GPT-4o-mini} & \textbf{GPT-4o} \\
        \midrule
        One Row  & 45.05 & 72.93 \\
        Multi Row  & 43.46 & 80.28 \\
        \bottomrule
    \end{tabular}
    \caption{Performance Comparison of Multi Row and One Row Schemas for GPT-4o-mini and GPT-4o on Calendar Scheduling}
    \label{tab:multi_vs_one}
\end{table}
\subsection{LLM  Struggles to Design Effective Schema for Complex Planning}
Unlike Calendar Scheduling, which focuses on selecting a single time slot, TravelPlanner involves generating a comprehensive travel itinerary, significantly increasing the complexity of the planning task. Our findings indicate that tasking the LLM with designing a table schema results in a notable performance drop compared to using direct prompting with a pre-defined schema. This suggests that LLMs currently lack the capability to independently design effective table schemas for complex planning tasks.

Although the provided schema is not perfect—omitting some critical columns, such as "cost" for budget constraints—it is generally more effective than most LLM-designed schemas. For instance, as shown in Table \ref{tab:given_vs_LLM_schema}, the LLM-developed schema and the given schema are structurally similar. However, a key difference is the use of "Dining Options" in the LLM-designed schema, as opposed to separating dining into specific categories like "breakfast," "lunch," and "dinner." In practice, this simplification often leads the LLM to allocate only a single meal per day, which contradicts commonsense expectations for travel planning.
\begin{table}[h]
    \centering
    \resizebox{0.5\textwidth}{!}{

    \begin{tabular}{lc}
        \toprule
        & \textbf{Schema Example} \\
        \midrule
        Given Schema  & days, current\_city, attraction, \\
        &  transportation, breakfast, \\
        &  lunch, dinner, accommodation  \\
        \midrule
        LLM Developed Schema & Day, Date, Location,   
 \\
        & Transportation Details,\\
        & Accommodation Details, \\
        & Activities/Attractions, Dining Options, \\
        & Estimated Cost, Notes/Preferences \\
        \bottomrule
    \end{tabular}
    }
    \caption{Given Schema and Example of GPT-4o developed Schema}
    \label{tab:given_vs_LLM_schema}
\end{table}

\subsection{Ablation Study}
\begin{table}[h!]
    \centering
    \resizebox{0.5\textwidth}{!}{
        \begin{tabular}{ccc}
            \toprule
            Schema Designing & Resoning Verification & ACC(\%) \\
            \midrule
            \checkmark & \checkmark & \textbf{42.3} \\
            \checkmark & \texttimes & 38.5 (\textcolor{red}{$\downarrow$\,3.8}) \\
    
            \texttimes & \checkmark & 36.2 (\textcolor{red}{$\downarrow$ 6.1}) \\
            \texttimes & \texttimes & 32.7 (\textcolor{red}{$\downarrow$ 9.6}) \\
            \bottomrule
        \end{tabular}
    }
    \caption{Ablation study results for GPT-4o-mini with schema designing and reasoning verification effects on performance of calendar scheduling.}
    \label{tab:ablation_study}
\end{table}
We conducted an ablation study using GPT-4o-mini on the Calendar Scheduling task to evaluate the individual contributions of schema design and reasoning verification . Table~\ref{tab:ablation_study} shows 
that when reasoning verification is removed, accuracy drops from 42.3\% to 38.5\% (\textcolor{red}{$\downarrow$ 3.8\%}). This indicates that without explicitly verifying constraints, the LLM may overlook key restrictions in the query, leading to false positives during self-checking. 
The absence of schema design leads to a larger performance drop, from 42.3\% to 36.2\% (\textcolor{red}{$\downarrow$ 6.1\%}), and further to 32.7\% (\textcolor{red}{$\downarrow$ 9.6\%}) when both schema design and reasoning verification are removed. This highlights the critical role of schema design in structuring the reasoning process.  Table~\ref{tab:w_wo_schema} shows that without a schema, the LLM tends to create tables with fewer columns, omitting key information necessary for constraint checking. While the table without schema design contains basic headers such as \texttt{Participant} and \texttt{Selected Meeting Time}, the schema-designed table includes additional headers like \texttt{Conflict Check}, \texttt{Work Hours Start/End}, and \texttt{Notes/Comments}. These additional columns capture critical reasoning steps and constraints, enabling more effective verification and selection of a valid meeting time.

\begin{table}[h]
    \centering
    \resizebox{0.5\textwidth}{!}{
    \begin{tabular}{lc}
        \toprule
        & \textbf{Column Headers} \\
        \midrule
        \textbf{wo/}  Schema Design  & Participant, Available Time Slots, \\
        & Selected Meeting Time \\
        \midrule
        \textbf{w/} Schema Design  & Participant Name, Participant Availability, \\
        & Meeting Duration, Meeting Day, \\ 
        & Proposed Meeting Time, Work Hours Start, \\
        &  Work Hours End,  Conflict Check, \\
        &  Final Meeting Time , Notes/Comments \\
        \bottomrule
    \end{tabular}
    }
    \caption{Example of Column Headers of Table Thoughts w/wo Schema Design }
    \label{tab:w_wo_schema}
\end{table}
\section{Conclusion}  
We proposed \AgentName, a novel framework that introduces structured reasoning at the thought level. The framework centers on the design and utilization of table schemas, where the LLM is tasked with constructing a schema and generating structured thoughts based on it. Our results demonstrate that \AgentName\ excels in constraint planning tasks, showcasing its ability to manage complex constraints effectively. Moreover, the framework exhibits significant potential for further improving performance in math reasoning tasks, particularly in addressing unsolved problems through structured reasoning.  

Additionally, we conducted detailed analyses of the results, exploring the interplay between schema design, reasoning complexity, and model capabilities. These insights pave the way for future research into the nature and representation of thought processes, offering a promising direction for the development of more robust reasoning frameworks in LLMs.

\section*{Limitations}  
Our proposed methods are currently supported only by models capable of generating structured data with complex schemas. This limitation restricts our experiments to a small set of closed-source models, such as those provided by OpenAI. Consequently, the generalizability of our findings to open-source LLMs remains unexplored. Future work should investigate approaches for adapting \AgentName\ to a broader range of models, including those with limited native support for structured data generation.

\section*{Ethical Statement}
This research was conducted using publicly available datasets (e.g., GSM-8K, MATH500, TravelPlanner) in compliance with their terms of use, ensuring no personally identifiable information (PII) was processed. While our proposed framework, \AgentName, aims to enhance structured reasoning in LLMs, we acknowledge the potential risks of misuse in harmful applications, such as deceptive planning or adversarial reasoning. To mitigate this, we advocate for responsible deployment with appropriate safeguards.

\bibliography{anthology,custom}
\appendix

\section{Completion Rate Analysis of \AgentName\ on Closed and Open Source LLMs}
\label{sec:open_fail}

We investigate the completion rate of \AgentName\ when using both closed-source and open-source LLMs in calendar scheduling tasks. Despite advancements in open-source models such as LLaMA 3.1-8B and LLaMA 3.1-70B, these models encounter significant challenges in adhering to the tool schema required for the schema design and table construction modules. Specifically, they fail to generate the expected outputs due to the complexity of the tool schema, leading to errors in the intermediate pipeline.

In a set of 100 test examples, we observe that \AgentName\ fails 100\% and 63\% of the time when using LLaMA 3.1-8B and LLaMA 3.1-70B, respectively. These high failure rates make experiments with open-source models infeasible because: 

\begin{enumerate}
    \item \textbf{High Computational Cost:} The need for repeated retries significantly increases computational overhead.
    \item \textbf{Result Bias:} Retrying may introduce bias, as incorrect reasoning processes may be verified as correct, prematurely terminating the pipeline and artificially increasing the success rate.
\end{enumerate}

In contrast, closed-source models such as GPT-4 variants demonstrate much higher success rates. The results are summarized in Table~\ref{tab:success_rates}.

\begin{table}[h!]
    \centering
    \caption{Success Rates of \AgentName\ with Different LLMs on Calendar Scheduling Tasks}
    \label{tab:success_rates}
    \begin{tabular}{lc}
        \toprule
        \textbf{Model} & \textbf{Completion Rate (\%)}  \\
        \midrule
        LLaMA 3.1-8B & 0  \\
        LLaMA 3.1-70B & 37  \\
        GPT-4o & 100 \\
        GPT-4o-mini & 99  \\
        \bottomrule
    \end{tabular}
\end{table}

\onecolumn
\section{Example of \AgentName\ on TraverPlanner Task}
\subsection{\AgentName\ w/ Given Schema}
\begin{tcolorbox}[colback=gray!10, colframe=gray!50, boxrule=0.5mm, sharp corners=southwest, breakable]
\textbf{Query:} Please draw up a 3-day travel itinerary for one person, beginning in Oakland and heading to Tucson from March 15th to March 17th, 2022, with a budget of \$1,400. 

\textbf{Raesoning table: }
\begin{lstlisting}
"day1": {
    "days": 1,
    "current_city": "Oakland",
    "transportation": "Flight Number: F4002752, from Oakland to Tucson, Departure Time: 15:07, Arrival Time: 17:00",
    "breakfast": "-",
    "attraction": "-",
    "lunch": "-",
    "dinner": "Pizza Street, Tucson",
    "accommodation": "Private room with private bathroom, Tucson",
    "total_cost": 270,
    "id": "day1"
},
"day2": {
    "days": 2,
    "current_city": "Tucson",
    "transportation": "-",
    "breakfast": "Mocha, Tucson",
    "attraction": "Pima Air & Space Museum, Tucson",
    "lunch": "Pizza Street, Tucson",
    "dinner": "Canteen Till I Die, Tucson",
    "accommodation": "Room for rent shared bathroom, Tucson",
    "total_cost": 61,
    "id": "day2"
},
"day3": {
    "days": 3,
    "current_city": "Tucson",
    "transportation": "Self-driving from Tucson to Oakland, Duration: 12 hours 42 mins, Cost: $68",
    "breakfast": "-",
    "attraction": "-",
    "lunch": "-",
    "dinner": "-",
    "accommodation": "-",
    "total_cost": 68,
    "id": "day3"
}
\end{lstlisting}
\end{tcolorbox}

\subsection{\AgentName\ w/ Self-designed Schema}
\begin{tcolorbox}[colback=gray!10, colframe=gray!50, boxrule=0.5mm, sharp corners=southwest, breakable]
\textbf{Query:} Please draw up a 3-day travel itinerary for one person, beginning in Oakland and heading to Tucson from March 15th to March 17th, 2022, with a budget of \$1,400. 

\textbf{Raesoning table: }

\begin{lstlisting}
"1": {
    "Day": "Day 1",
    "Date": "2022-03-15",
    "Travel Mode": "Flight",
    "Departure Location": "Oakland",
    "Destination": "Tucson",
    "Travel Duration": "1 hour 53 minutes",
    "Travel Cost": 190,
    "Accommodation Name": "Private room with private bathroom",
    "Accommodation Cost": 58,
    "Room Type": "Private room",
    "House Rules": "No smoking",
    "Attractions": "Pima Air & Space Museum (Cost: $15), Reid Park Zoo (Cost: $10)",
    "Dining Options": "Villa Tevere (Cost: $37), Magic Spice Wok (Cost: $31)",
    "Total Trip Cost": 341,
    "Budget Remaining": 1059,
    "id": "1",
    "Maximum Occupancy": "2 people"
},
"2": {
    "Day": "Day 2",
    "Date": "2022-03-16",
    "Travel Mode": "Self-driving",
    "Departure Location": "Tucson",
    "Destination": "Oakland",
    "Travel Duration": "12 hours 40 minutes",
    "Travel Cost": 68,
    "Accommodation Name": "Private room with private bathroom",
    "Accommodation Cost": 58,
    "Room Type": "Private room",
    "House Rules": "No smoking",
    "Attractions": "Tucson Botanical Gardens (Cost: $15), Old Tucson (Cost: $20)",
    "Dining Options": "La Plage (Cost: $93), Ooh Lala! (Cost: $70)",
    "Total Trip Cost": 324,
    "Budget Remaining": 735,
    "id": "2",
    "Maximum Occupancy": "2 people"
},
"3": {
    "Day": "Day 3",
    "Date": "2022-03-17",
    "Travel Mode": "Self-driving",
    "Departure Location": "Tucson",
    "Destination": "Oakland",
    "Travel Duration": "12 hours 42 minutes",
    "Travel Cost": 68,
    "Accommodation Name": "None",
    "Accommodation Cost": 0,
    "Room Type": "Private room",
    "House Rules": "No smoking",
    "Attractions": "Arizona-Sonora Desert Museum (Cost: $15), Children's Museum Tucson (Cost: $10)",
    "Dining Options": "Pirates of Grill (Cost: $52), Mood 4 Food (Cost: $20)",
    "Total Trip Cost": 165,
    "Budget Remaining": 570,
    "id": "3",
    "Maximum Occupancy": "N/A"
}
\end{lstlisting}
\end{tcolorbox}

\section{Example of \AgentName\ on Calendar Scheduling Task}

\subsection{\AgentName\ w/ One Row Schema}
\begin{tcolorbox}[colback=gray!10, colframe=gray!50, boxrule=0.5mm, sharp corners=southwest, breakable]
\textbf{Query:} You need to schedule a meeting for Roy, Kathryn and Amy for half an hour between the work hours of 9:00 to 17:00 on Monday. 

Here are the existing schedules for everyone during the day: Roy has blocked their calendar on Monday during 9:00 to 9:30, 10:00 to 10:30, 11:00 to 11:30, 12:30 to 13:00; Kathryn has meetings on Monday during 9:30 to 10:00, 16:30 to 17:00; 
Amy has blocked their calendar on Monday during 9:00 to 14:30, 15:00 to 16:00, 16:30 to 17:00; 

Amy would rather not meet on Monday after 15:30. Find a time that works for everyone's schedule and constraints.

\textbf{Raesoning table: }

\begin{lstlisting}
 "1": {
    "Meeting Duration": "30 minutes",
    "Work Hours Start": "9:00",
    "Work Hours End": "17:00",
    "Roy's Unavailable Times": "9:00-9:30, 10:00-10:30, 11:00-11:30, 12:30-13:00",
    "Kathryn's Unavailable Times": "9:30-10:00, 16:30-17:00",
    "Amy's Unavailable Times": "9:00-14:30, 15:00-16:00, 16:30-17:00",
    "Amy's Preference": "Prefers not to meet after 15:30",
    "Proposed Meeting Time": "14:30-15:00"
}
\end{lstlisting}
\end{tcolorbox}
\subsection{\AgentName\ w/ Multi Row Schema}
\begin{tcolorbox}[colback=gray!10, colframe=gray!50, boxrule=0.5mm, sharp corners=southwest, breakable]
\textbf{Query:} You need to schedule a meeting for Kathryn, Charlotte and Lauren for half an hour between the work hours of 9:00 to 17:00 on Monday. 

Here are the existing schedules for everyone during the day: 
Kathryn has blocked their calendar on Monday during 9:00 to 9:30, 10:30 to 11:00, 11:30 to 12:00, 13:30 to 14:30, 16:30 to 17:00; 
Charlotte has blocked their calendar on Monday during 12:00 to 12:30, 16:00 to 16:30; 
Lauren has blocked their calendar on Monday during 9:00 to 10:00, 12:00 to 12:30, 13:30 to 14:30, 15:00 to 16:00, 16:30 to 17:00;

Charlotte do not want to meet on Monday after 13:30. Find a time that works for everyone's schedule and constraints.

\textbf{Raesoning table: }

\begin{lstlisting}
"1": {
    "Participant Name": "Kathryn",
    "Unavailable Time Slots": "9:00-9:30, 10:30-11:00, 11:30-12:00, 13:30-14:30, 16:30-17:00",
    "Preferred Meeting Time": "None",
    "Meeting Duration": "30 minutes",
    "Work Hours": "9:00-17:00",
    "Available Time Slots": "9:30-10:30, 11:00-11:30, 12:00-13:30, 14:30-16:30",
    "Constraints Satisfied": true,
    "Proposed Meeting Time": "10:00-10:30"
},
"2": {
    "Participant Name": "Charlotte",
    "Unavailable Time Slots": "12:00-12:30, 16:00-16:30",
    "Preferred Meeting Time": "Before 13:30",
    "Meeting Duration": "30 minutes",
    "Work Hours": "9:00-17:00",
    "Available Time Slots": "9:00-12:00, 12:30-16:00, 16:30-17:00",
    "Constraints Satisfied": true,
    "Proposed Meeting Time": "10:00-10:30"
},
"3": {
    "Participant Name": "Lauren",
    "Unavailable Time Slots": "9:00-10:00, 12:00-12:30, 13:30-14:30, 15:00-16:00, 16:30-17:00",
    "Preferred Meeting Time": "None",
    "Meeting Duration": "30 minutes",
    "Work Hours": "9:00-17:00",
    "Available Time Slots": "10:00-12:00, 12:30-13:30, 14:30-15:00",
    "Constraints Satisfied": true,
    "Proposed Meeting Time": "10:00-10:30"
}
\end{lstlisting}
\end{tcolorbox}
\section{Comparison of \AgentName\ and Direct Prompting on GSM-8k}
\subsection{Example where Direct Prompting  succeceds while \AgentName\ Fails}
\begin{tcolorbox}[colback=gray!10, colframe=gray!50, boxrule=0.5mm, sharp corners=southwest, breakable]
\textbf{Query:} Stephen placed an online order for groceries.  His final bill came to \$40.00.  Because this was through a delivery vendor, they tacked on a 25\% fee to his final total and charged him \$3.00 in delivery fees.  Stephen also added a \$4.00 tip.  After the extra fees, what was the final price of Stephen's groceries?

\textbf{Raesoning table: }

\begin{lstlisting}
"1": {
    "Base_Price": 40.0,
    "Delivery_Fee_Percentage": 0,
    "Delivery_Fee_Amount": 0,
    "Tip_Amount": 0,
    "Total_Fee_Percentage": 0,
    "Total_Fee_Amount": 0,
    "Final_Price": 40.0
}
\end{lstlisting}

\textbf{Direct Prompting response:} 

To find the final price of Stephen's groceries after all the extra fees, we need to calculate each component and add them together.

1. \textbf{Original Bill}: Stephen's original grocery bill was \$40.00.

2. \textbf{Vendor Fee}: The vendor adds a 25\% fee to the original bill. To calculate this fee:
\[
\text{Vendor Fee} = 0.25 \times 40.00 = 10.00
\]

3. \textbf{Delivery Fee}: The delivery fee is a flat \$3.00.

4. \textbf{Tip}: Stephen added a \$4.00 tip.

Now, we add all these amounts together to find the final price:
\[
\text{Final Price} = \text{Original Bill} + \text{Vendor Fee} + \text{Delivery Fee} + \text{Tip}
\]
\[
\text{Final Price} = 40.00 + 10.00 + 3.00 + 4.00 = 57.00
\]

Therefore, the final price of Stephen's groceries, after all the extra fees, is \$57.00.

\end{tcolorbox}
\subsection{Example where \AgentName\ succeceds while Direct Prompting Fails}
\begin{tcolorbox}[colback=gray!10, colframe=gray!50, boxrule=0.5mm, sharp corners=southwest, breakable]
\textbf{Query:} Jamal's phone can hold 6 times more photographs than can Brittany's phone.  The maximum number of photographs that Brittany's phone can hold is 50 times more than the number of birds in Jamal's photograph of the ducks at the zoo.  If Jamal's phone can hold 1800 photographs, how many ducks can be seen in Jamal's photograph of ducks at the zoo?

\textbf{Raesoning table: }

\begin{lstlisting}
"1": {
    "Jamal_Phone_Capacity": 1800,
    "Brittany_Phone_Capacity": 300,
    "Ducks_in_Photograph": 6,
    "Brittany_Phone_Additional_Capacity": 50,
    "Equation": "300 = 50D",
    "Solution_Steps": "1. Jamal's phone can hold 6 times more photographs than Brittany's phone. 2. Jamal's phone can hold 1800 photographs. 3. Let B be the number of photographs Brittany's phone can hold. Then, 6B = 1800. 4. Solve for B: B = 1800 / 6 = 300. 5. Brittany's phone can hold 50 times more photographs than the number of ducks in Jamal's photograph. 6. Let D be the number of ducks in Jamal's photograph. B = 50D. 7. Substitute B = 300 into the equation: 300 = 50D. 8. Solve for D: D = 300 / 50 = 6.",
    "Final_Answer": 6
}
\end{lstlisting}

\textbf{Direct Prompting response:} 

To solve the problem, we need to determine the number of ducks in Jamal's photograph.

First, let's define the variables:
\begin{itemize}
    \item Let \( B \) be the maximum number of photographs Brittany's phone can hold.
    \item Let \( D \) be the number of ducks in Jamal's photograph.
\end{itemize}

According to the problem, Jamal's phone can hold 6 times more photographs than Brittany's phone. Therefore, we have the equation:
\[
J = 6B
\]
where \( J \) is the number of photographs Jamal's phone can hold. We know from the problem that \( J = 1800 \). Substituting this into the equation gives:
\[
1800 = 6B
\]

To find \( B \), divide both sides by 6:
\[
B = \frac{1800}{6} = 300
\]

Next, the problem states that Brittany's phone can hold 50 more photographs than the number of ducks in Jamal's photograph. This gives us the equation:
\[
B = D + 50
\]

We already found that \( B = 300 \). Substituting this into the equation gives:
\[
300 = D + 50
\]

To find \( D \), subtract 50 from both sides:
\[
D = 300 - 50 = 250
\]

Thus, the number of ducks in Jamal's photograph is \( \boxed{250} \).

\end{tcolorbox}

\end{document}